\documentclass[a4paper,conference]{IEEEtran}
\ifCLASSINFOpdf
\else
\fi
\usepackage[pdftex]{graphicx}
\graphicspath{./}
\DeclareGraphicsExtensions{.png}
\usepackage{amsmath}
\usepackage{url}

\begin{document}

\title{Network Comparison Study of\\ Deep Activation Feature Discriminability\\ with Novel Objects}

\author{\IEEEauthorblockN{Michael Karnes}
\IEEEauthorblockA{Department of Civil, \\Environmental and Geodetic Engineering\\
The Ohio State University\\
Columbus, Ohio 43210\\
Email: karnes.30@osu.edu }
\and
\IEEEauthorblockN{Alper Yilmaz}
\IEEEauthorblockA{Department of Civil, \\Environmental and Geodetic Engineering\\
The Ohio State University\\
Columbus, Ohio 43210\\
Email: yilmaz.15@osu.edu}}

\maketitle

\begin{abstract}
Feature extraction has always been a critical component of the computer vision field. More recently, state-of-the-art computer visions algorithms have incorporated Deep Neural Networks (DNN) in feature extracting roles, creating Deep Convolutional Activation Features (DeCAF). The transferability of DNN knowledge domains has enabled the wide use of pretrained DNN feature extraction for applications with novel object classes, especially those with limited training data. This study analyzes the general discriminability of novel object visual appearances encoded into the DeCAF space of six of the leading visual recognition DNN architectures. The results of this study characterize the Mahalanobis distances and cosine similarities between DeCAF object manifolds across two visual object tracking benchmark data sets. The backgrounds surrounding each object are also included as an object classes in the manifold analysis, providing a wider range of novel classes. This study found that different network architectures led to different network feature focuses that must to be considered in the network selection process. These results are generated from the VOT2015 and UAV123 benchmark data sets; however, the proposed methods can be applied to efficiently compare estimated network performance characteristics for any labeled visual data set.
\end{abstract}


%
\IEEEpeerreviewmaketitle

\section{Introduction}
Deep neural networks (DNN) provide flexible function structures for modeling high dimensional patterns making them highly effective for image processing applications \cite{ILSVRC15,VOT_TPAMI,PR_review_2018,zhao2019object}. Since their origination with LeNet \cite{LeCun89}, DNN have been viewed as nested feature embedding functions  that condense the high dimensional image space into a lower dimensioned deep convolutional activation feature (DeCAF) space \cite{donahue2014decaf,zhu2019} on which the final classification decision is made. 

\begin{align}
y=f_{n}(...f_{3}(f_{2}(f_{1}(x)))
\end{align} 

Other feature extraction methods, such as HoG, SIFT, and ORB, are also used for a similar information condensation process, representing image regions as descriptors in the feature space \cite{Levine_1969,karami-2017,Humeau-Heurtier-2019,HoG2005,SIFT2005,orb2011}. They were first developed for keypoint matching in visual mapping \cite{Krig2016}; and now have spread to a variety of visual tasks including image classification \cite{pandey2015passive}, object recognition \cite{kortli,ali2016}, and object localization \cite{monzo2011precise}. There have been continual developments in feature design to improve computational efficiency, view invariance, and object discriminability \cite{feature_survey_2018}. 

The primary advantage to using pretrained DNN is the reduction in training requirements. This approach assumes that the novel task has a similar knowledge domain to the trained as seen in incremental learning, network fine tuning, transfer learning, and feature encoding \cite{pan2009survey,zheng2017sift}. Without the loss of generality, our work focuses on feature encoding for long-term tracking scenarios. Long-term tracking is a strong choice for studying the DNN activation manifolds of novel targets for two reasons. The first is the availability of trusted benchmark data sets of novel objects. The second being the object's variable appearances through a sequence. On local time scales, such as 10 frames, the changes in the appearance of the object are minor. As the sequence progresses, the range of objects appearances increases providing samples of object appearances in different positions and from different camera perspectives.

The DNN ability to generate highly descriptive complex filters affords DeCAF encoding a distinct advantage. This work aims to characterize the encoded spaces of the most prevalent image recognition network architectures and provide a methodology for quantitatively measuring the relative difficulty in learning a set of novel targets. Better understanding of DeCAF behaviors enables more informed network selections.

We developed this DeCAF characterization methodology to help our team select the best network for DeCAF encoding for a particular data set. This is precisely where our contributions are aimed. In this study:
\begin{itemize}

\item We propose a novel methodology for characterizing the object manifold discriminability across any annotated custom data set.
\item We present the first generalized DeCAF manifold discriminability characterization. This comparative DeCAF survey analyzed the DeCAF manifolds of 294 novel classes over six top image recognition network architectures.
\item We provide novel results that demonstrate the differences in learned features between network architectures.
\end{itemize}

\section{Related Work}
AlexNet changed the computer vision field with its outstanding performance in the 2012 ImageNet object recognition competition \cite{alex_2012}. One of these changes was the resurgence in study of trained network knowledge transferability to novel classification, detection, and tracking tasks. The deep activation features (DeCAF) provided a new means of efficiently encoding the intricate patterns of object appearances. Applying DeCAF to novel tasks takes the same general form of few-shot learning with the goal of transforming the raw data space to a discriminative feature space that maximizes the manifold model accuracy and separability with limited training samples\cite{Wang_2020}.

The developments in generalized tracking are a great example of the successful incorporation DeCAF. The constraints of the tracking scenario provide a well defined scope for few-shot learning. Prior to the creation of DeCAF, discriminative correlation filters (DCF) used HoG and color features to generate object bounding masks \cite{bolme2010visual}. The incorporation of the DeCAF further extended the success of DCF producing a line of award winning trackers \cite{Danelljan2014,Danelljan2015,Danelljan2016,Danelljan2016_2,Danelljan2017eco}.

The classification problem is more ambiguous than the tracking. With tracking, the training samples and the potential object appearances are temporially limited. The constraints around the classification problem are less well defined. In single object tracking, there are only the object and background manifolds to separate and the manifolds are limited to the potential appearance in the next frame.  In classification, the manifold of each object proportionally increases with potential appearance variations and the number of manifolds increases to the number of considered objects. As a result, classification algorithms replace the based DCF model with computationally simpler similarity metrics, such as euclidean distances, cosine similarities, SVM, and logistic regression \cite{Wang_2020}.

The application of DeCAF in classification coincided with its incorporation into generalized tracking. \cite{donahue14} was the first to analyze the separability of novel DeCAF manifolds across layers of the AlexNet finding the best discrimination in deeper layers and a significant improvement over SURF. Developments then moved toward applying DeCAF to zero-shot learning \cite{Snell2017} with comparisons between AlexNet, VGGNet, and GoogLeNet. Now DeCAF few-shot learning has extended into other domains such as remote sensing and medical image processing \cite{Mateen2019,RM2019,Xu_2017,Liu_2017}.

The exploration of the DeCAF space behavior has followed its rise in usage. The earliest characterization of DeCAF space focused on the effect of layer choice in the AlexNet for generalized classification \cite{Ma_2015_ICCV}. Many other studies have followed similar suite, investigating the effect of network and layer selection on the performance of their algorithm\cite{Ma_2015_ICCV,wang2018face,wang-2018}. Recent studies have begun to focus on characterizing DeCAF behaviors in a more generalized manner. One example of such study investigated the affine mappings between GoogLeNet and ResNet DeCAF manifolds \cite{Mcneely2020}. Another interesting example investigated the theoretical classification capacity of the AlexNet and VGGNet across layers \cite{cohen_2020}. 

Our study adds a practice driven approach to this line of generalized DeCAF behavior investigations. The high cost of brute-force searching limits the number networks considered in the selection process. The manifold estimation methodologies of few-shot learning provide scalable metrics well suited for empirically characterizing object discriminability across several DeCAF spaces.

\section{Methods}
\subsection{Overview}
The DeCAF characterization process has three steps: preprocessing images, encoding the image sets into the DeCAF space, and analyzing class DeCAF manifolds.

The DeCAF manifolds are extracted from the deepest convolutional layers of the six top performing image recognition DNN architectures trained on the ILSVRC\cite{ILSVRC15} data set: VGG19, InceptionV3, ResNet50, DenseNet121, MobileNet, NASNetLarge. These networks were carefully selected to represent the evolutionary line of image recognition network architectures. 

\subsection{Sampling}
The DeCAF manifolds are generated by encoding the image space containing the object annotated bounding boxes for each frame in the sample set with each network. The resulting generated data set consists of the means and variances of each DeCAF manifold for the target object and background in each frame. It is these distribution metrics that are then used to analyze the separability and similarity between classes.

Two data sets are used to generate DeCAF manifolds in these analyses: VOT2015 and UAV123. The VOT2015 was selected due to its intentional design to represent a wide range of visual tracking scenarios. It contains 60 sequences with a variety of classes and scenes. The UAV123 was selected to represent an alternative data set. It contains 87 sequences from a UAV point of view. The classes in this set focus mostly on road scenes. In total, 147 unique sequences are included, creating a total of 294 novel classes.

\subsection{Metrics}
Metric selection was a serious consideration for this study. Several candidates were considered: SVM, Logistic regression, KL divergence, Mahalanobis distance, and cosine similarity. In the end, the Mahalanobis distance, and cosine similarity metrics are selected to quantify manifold discriminability. This decision was heavily driven by their low computational cost and robustness in low rank, high dimensional spaces \cite{mohapatra_2014}. These metrics have the distinct advantage of analytically updated models. This with their strong performance make them popular choices in few-shot learning scenarios \cite{Wang_2020}.

The cosine similarity is calculated by the dot product of the normalized manifold centroids.

\begin{align}
\theta &= \langle x_1, x_2\rangle = \underset{i} \Sigma x_{1i} x_{2i}
\end{align}  

The Mahalanobis distance is the covariance standardized Euclidean distance between manifold centroids. This calculation is simplified by assuming dimensional independence which allows for the covariance to be estimated as a diagonal matrix.

\begin{align}
\mathbf{Q}=diag(\sigma_i^2)\\
d=(x_1-x_2)\mathbf{Q}^{-1}(x_1-x_2)^{T}
\end{align}

\section{Experiments}
This set of experiments looks to answer three questions:
1.) How does class discriminability vary across network architectures?
2.) How does the sample set, (number of frames, ordered vs random sampling), process effect class manifolds?
3.) Are the network behaviors extendable to other data sets?

These experiments begin analyzing the target object (TG) and background (BG) manifolds across the entire 60 VOT2015 classes over the six considered networks. Then the effects of sampling are investigated, specifically the effects of sample set size and sampling noise. The final experiment investigates the consistency of DeCAF behavior across data sets with the UAV123.

\section{Results}
\subsection{Plot Interpretation}
The presentation of these results can seem a overly complex at first glance; please refer to \ref{fig:VOT_Full}. Be assured that the formatting is held consistent through the study. The plots are designed to enable the direct comparison of metaclasses TG and BG similarities across all networks. This has led to the cell and matrix format. The plot contains six cells, one for each of the networks arranged in two rows of three. Each cell contains a two by two matrix showing the target object and background similarities for that network. The diagonals of the matrix show the within metaclass mean TG-TG and BG-BG similarities pooled across all classes. The off diagonals show the TG-BG similarities pooled across all object classes. This plot structure is used to present both the cosine similarity, shown in grey scale, and Mahalanobis distance, shown in color.

\subsection{Full VOT2015}
\begin{figure}
  \centering
  \includegraphics[width=0.7\linewidth]{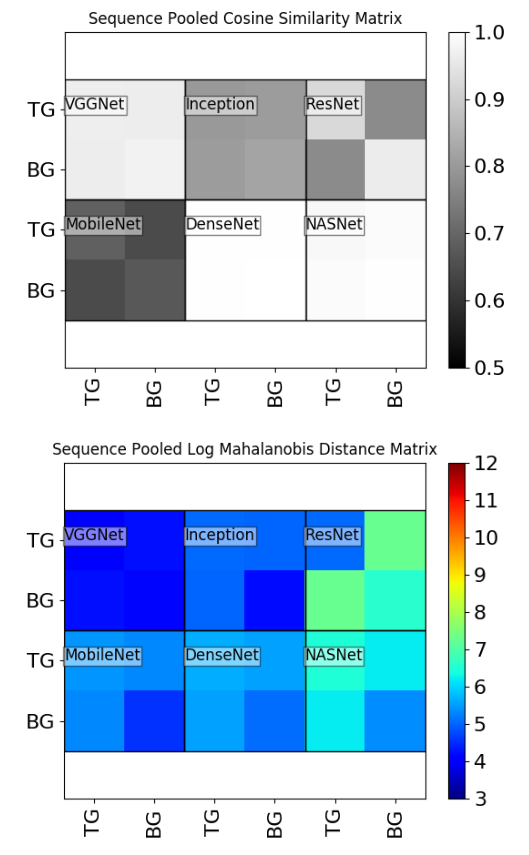}
  \caption{Top: This plot shows the mean sequence pooled cosine similarities for the sequences in the VOT2015 data set. Bottom: This plot shows the log mean sequence pooled Mahalanobis distances for the sequences in the VOT2015 data set. }
  \label{fig:VOT_Full}
\end{figure}

This experimental case analyzes the DeCAF manifolds of the 60 full sequences in the VOT2015 data set. These manifolds contain the appearances of the target objects and their backgrounds seen throughout the sequence. The similarities of these manifolds are presented in Figure \ref{fig:VOT_Full}. At first glance, it is obvious that the networks behave differently with target object and background encoding. For example, the Mahalanobis distance plot of the ResNet shows a uniquely high discriminability of the background manifolds. This is especially interesting while also considering its relatively low TG-TG separation.

The top performing DeCAF space varied with task. The MobileNet had the lowest cosine similarities across all classes. The NASNet had the largest Mahalanobis distance for TG-TG discrimination. The ResNet had the largest TG-BG distance.

\subsection{VOT2015 Set Size}
\begin{figure}
  \centering
  \includegraphics[width=0.7\linewidth]{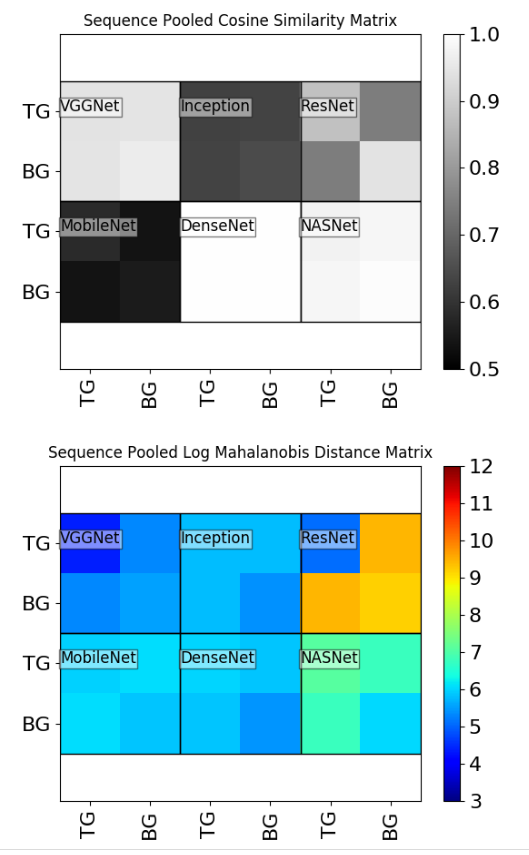}
  \caption{DeCAF characterization of the six considered networks with 10 sequential frames from each sequence in VOT2015 data set.}
  \label{fig:VOT_10}
\end{figure}

\begin{figure}
  \centering
  \includegraphics[width=0.7\linewidth]{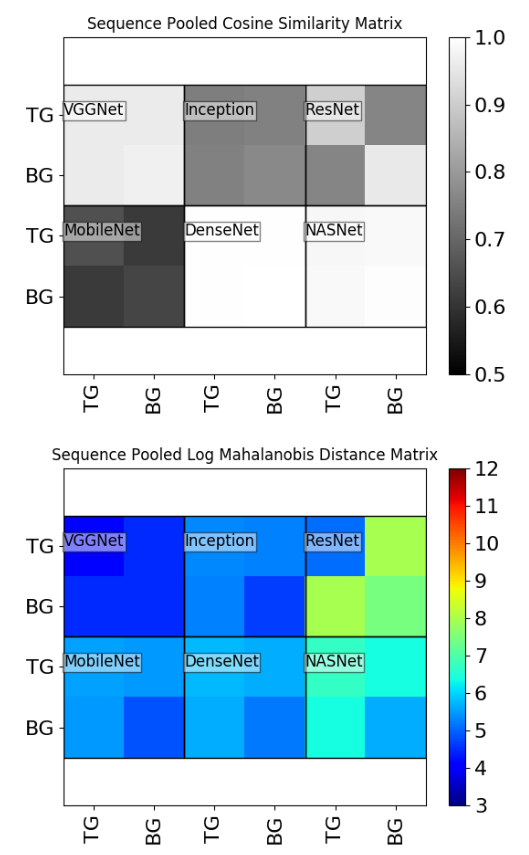}
  \caption{DeCAF characterization of the six considered networks with 100 sequential frames from each sequence in VOT 2015.}
  \label{fig:VOT_100}
\end{figure}

This experimental case investigates the effect of sample set size on manifold separability. This was done by analyzing the manifolds generated from sets of 10 and 100 sequential frames. The results for these analyses are presented in Figures \ref{fig:VOT_10} and \ref{fig:VOT_100}.

The primary effect of smaller sample set sizes was an increase in class manifold discriminability. This effect is visible in both the 10 frame and 100 frame cases. The greatest effect was seen in the ResNet BG-BG Mahalanobis distances. Restricting the sample set to 10 frames raised the BG-BG distance magnitude to \(10^9\). With 100 frames it was \(10^8\) and with the full sequences it was \(10^7\). The cosine similarities were far less impacted showing relatively small changes in discriminability.

Sample set size also had effects on network behavior. This is seen the changes in the TG-TG to BG-BG Mahalanobis distance ratio of the VGGNet. In the 10 frame case, the BG-BG distance is larger than the TG-TG. As the set size increases to 100 frame and full sequences, the differences in these distances decrease, suggesting a change in relative knowledge domain with changes in sample set size.


\subsection{VOT2015 Random Sampling}
\begin{figure}
  \centering
  \includegraphics[width=0.7\linewidth]{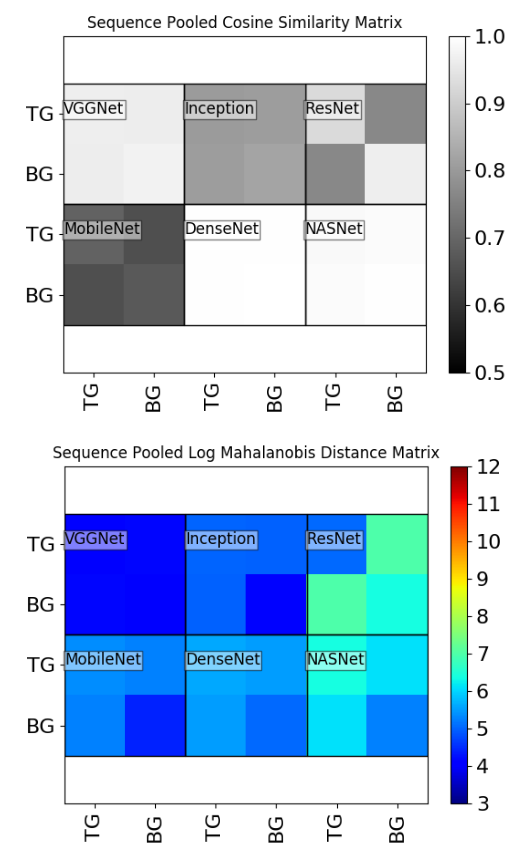}
  \caption{DeCAF characterization of the six considered networks with 1000 frame sets randomly sampled from each sequence in VOT2015 with an induced bounding box noise of three pixels.}
  \label{fig:VOT_Rand}
\end{figure}

This experimental case investigates the effect of noise in the sampling process. This was achieved by randomly sampling a 1000 frames from each sequence in the VOT2015 and inducing a three pixel noise to the position and size of the annotated bounding boxes. The results of this experiment are presented in Figure \ref{fig:VOT_Rand}. The introduction of noise had a negligible impact on manifold separation in both cosine similarities in Mahalanobis distances with only a slight trend in decreased discriminability seen upon close inspection. The most notable changes occurred with the ResNet and NASNet relative TG-BG separation. In both cases, the introduction of noise caused a relative decrease in the TG-BG Mahalanobis distances.



\subsection{Full UAV123}
\begin{figure}
  \centering
  \includegraphics[width=0.7\linewidth]{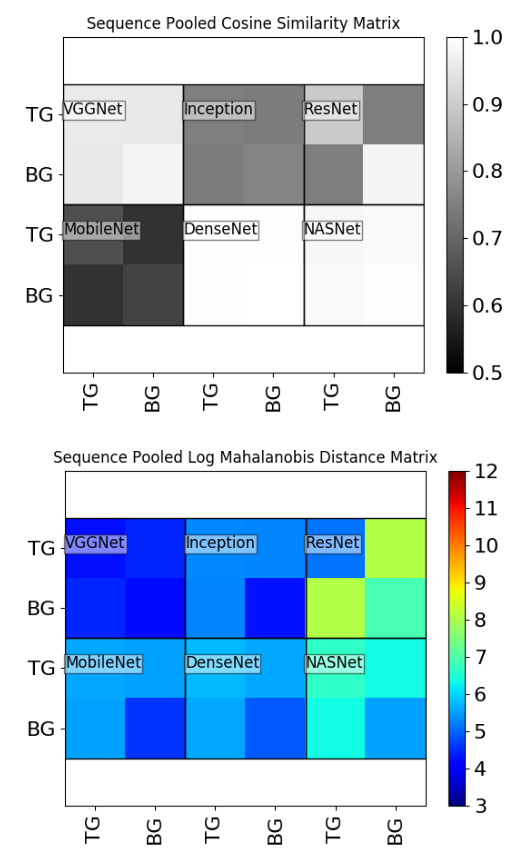}
  \caption{DeCAF characterization of the six considered networks on the full UAV123.}
  \label{fig:UAV_Full}
\end{figure}

This experimental case investigates the consistency of DeCAF behavior across data sets by analyzing the DeCAF manifolds of the full UAV123 data set. The results of this analysis are presented in Figure \ref{fig:UAV_Full}. The DeCAF behaviors remained consistent with those seen in with the VOT2015. The MobileNet provided the lowest cosine similarities across classes. The NASNet produced the largest Mahalanobis distances between TG-TG manifolds and the ResNet produced the largest TG-BG separations.

\section{Discussion}
The primary objective of these experiments is to investigate the generalized DeCAF novel object discriminability for network selection in few-shot learning applications. These experiments investigated the effects of sample set size, sampling noise, and data set. Discriminability of the encoded class manifolds was estimated with cosine similarities and Mahalanobis distances. The network DeCAF behaviors remained consistent across cases. In all cases, the MobileNet had the lowest cosine similarities; the NASNet had the largest TG-TG Mahalanobis distances; and the ResNet had the largest TG-BG Mahalanobis distances.

General shifts in manifold Mahalanobis distances were seen with changes in the sample set. The cosine similarities was less effected by sample set changes. Reducing the sample set size to 10 and 100 frames increased the Mahalanobis distance by factors of 10. Adding noise to the sampling process causes small decrease in manifold distances.

The most striking finding of this study is the diversity in DeCAF behaviors across different networks. The ResNet is unique in its focus on background classes. In all cases, the BG-BG Mahalanobis distance of the ResNet was larger than the TG-TG. All other networks displayed the opposite behavior with a greater focus of TG-TG discrimination with the exception of the VGGNet. In four of the five cases, the VGGNet showed equal focus on TG-TG and BG-BG discrimination. In the 10 frame case, the VGGNet showed a  BG-BG focused behavior similar to the ResNet.

Unique network behaviors were also seen in the cosine similarities. VGGNet, DenseNet, and NASNet showed high class similarities across all classes and cases. The ResNet and MobileNet displayed higher cosine similarities within meta-classes (TG-TG and BG-BG) and lower similarities for the TG-BG meta-class. The MobileNet produced the lowest cosine similarities across all cases and classes.

\section{Conclusions}
Deep convolutional activation features (DeCAF) are an efficient way of systematically extending the knowledge domain of pretrained DNN. This study characterized the DeCAF space of six of the leading image recognition architectures across 294 novel classes investigating the effects of sample set size, sampling noise, and data sets. The DeCAF spaces were characterized by the cosine similarities and Mahalanobis distances. It was found that network architectures have consistent behavior across cases. The cosine similarity showed lower class discrimination compared to Mahalanobis distance but was less effected by sampling changes. The NASNet produced the most discriminative features for object recognition. The ResNet produced the most discriminative features for object detection.

The most striking finding of this study was the distinct behaviors of each network illustrated by the varying focus on object versus background features. These findings clearly demonstrated the need for task appropriate network selections. The results of this study suggest consistent behaviors. None the less, researchers are most interested in network responses to their data sets. The proposed DeCAF characterization method can be applied to any annotated image set to efficiently estimate the discriminability of their DeCAF manifolds.

\bibliographystyle{./IEEEtran} 
\bibliography{main}

\begin{thebibliography}{10}
\providecommand{\url}[1]{#1}
\csname url@samestyle\endcsname
\providecommand{\newblock}{\relax}
\providecommand{\bibinfo}[2]{#2}
\providecommand{\BIBentrySTDinterwordspacing}{\spaceskip=0pt\relax}
\providecommand{\BIBentryALTinterwordstretchfactor}{4}
\providecommand{\BIBentryALTinterwordspacing}{\spaceskip=\fontdimen2\font plus
\BIBentryALTinterwordstretchfactor\fontdimen3\font minus
  \fontdimen4\font\relax}
\providecommand{\BIBforeignlanguage}[2]{{%
\expandafter\ifx\csname l@#1\endcsname\relax
\typeout{** WARNING: IEEEtran.bst: No hyphenation pattern has been}%
\typeout{** loaded for the language `#1'. Using the pattern for}%
\typeout{** the default language instead.}%
\else
\language=\csname l@#1\endcsname
\fi
#2}}
\providecommand{\BIBdecl}{\relax}
\BIBdecl

\bibitem{ILSVRC15}
O.~Russakovsky, J.~Deng, H.~Su, J.~Krause, S.~Satheesh, S.~Ma, Z.~Huang,
  A.~Karpathy, A.~Khosla, M.~Bernstein, A.~C. Berg, and L.~Fei-Fei, ``{ImageNet
  Large Scale Visual Recognition Challenge},'' \emph{International Journal of
  Computer Vision (IJCV)}, vol. 115, no.~3, pp. 211--252, 2015.

\bibitem{VOT_TPAMI}
M.~Kristan, J.~Matas, A.~Leonardis, T.~Vojir, R.~Pflugfelder, G.~Fernandez,
  G.~Nebehay, F.~Porikli, and L.~\v{C}ehovin, ``A novel performance evaluation
  methodology for single-target trackers,'' \emph{IEEE Transactions on Pattern
  Analysis and Machine Intelligence}, vol.~38, no.~11, pp. 2137--2155, Nov
  2016.

\bibitem{PR_review_2018}
\BIBentryALTinterwordspacing
D.~Bhamare and P.~Suryawanshi, ``Review on reliable pattern recognition with
  machine learning techniques,'' \emph{Fuzzy Information and Engineering},
  vol.~10, no.~3, pp. 362--377, 2018. [Online]. Available:
  \url{https://doi.org/10.1080/16168658.2019.1611030}
\BIBentrySTDinterwordspacing

\bibitem{zhao2019object}
Z.-Q. Zhao, P.~Zheng, S.-t. Xu, and X.~Wu, ``Object detection with deep
  learning: A review,'' \emph{IEEE transactions on neural networks and learning
  systems}, vol.~30, no.~11, pp. 3212--3232, 2019.

\bibitem{LeCun89}
\BIBentryALTinterwordspacing
Y.~LeCun, B.~Boser, J.~S. Denker, D.~Henderson, R.~E. Howard, W.~Hubbard, and
  L.~D. Jackel, ``Backpropagation applied to handwritten zip code
  recognition,'' \emph{Neural Computation}, vol.~1, no.~4, pp. 541--551, 1989.
  [Online]. Available: \url{https://doi.org/10.1162/neco.1989.1.4.541}
\BIBentrySTDinterwordspacing

\bibitem{donahue2014decaf}
J.~Donahue, Y.~Jia, O.~Vinyals, J.~Hoffman, N.~Zhang, E.~Tzeng, and T.~Darrell,
  ``Decaf: A deep convolutional activation feature for generic visual
  recognition,'' in \emph{International conference on machine learning}, 2014,
  pp. 647--655.

\bibitem{zhu2019}
\BIBentryALTinterwordspacing
J.~Zhu, H.~Yang, N.~Liu, M.~Kim, W.~Zhang, and M.~Yang, ``Online multi-object
  tracking with dual matching attention networks,'' \emph{CoRR}, vol.
  abs/1902.00749, 2019. [Online]. Available:
  \url{http://arxiv.org/abs/1902.00749}
\BIBentrySTDinterwordspacing

\bibitem{Levine_1969}
M.~D. {Levine}, ``Feature extraction: A survey,'' \emph{Proceedings of the
  IEEE}, vol.~57, no.~8, pp. 1391--1407, Aug 1969.

\bibitem{karami-2017}
E.~Karami, S.~Prasad, and M.~Shehata, ``Image matching using sift, surf, brief
  and orb: Performance comparison for distorted images,'' 2017.

\bibitem{Humeau-Heurtier-2019}
A.~{Humeau-Heurtier}, ``Texture feature extraction methods: A survey,''
  \emph{IEEE Access}, vol.~7, pp. 8975--9000, 2019.

\bibitem{HoG2005}
N.~{Dalal} and B.~{Triggs}, ``Histograms of oriented gradients for human
  detection,'' in \emph{2005 IEEE Computer Society Conference on Computer
  Vision and Pattern Recognition (CVPR'05)}, vol.~1, June 2005, pp. 886--893
  vol. 1.

\bibitem{SIFT2005}
\BIBentryALTinterwordspacing
D.~G. Lowe, ``Distinctive image features from scale-invariant keypoints,''
  \emph{Int. J. Comput. Vision}, vol.~60, no.~2, p. 91–110, Nov. 2004.
  [Online]. Available: \url{https://doi.org/10.1023/B:VISI.0000029664.99615.94}
\BIBentrySTDinterwordspacing

\bibitem{orb2011}
E.~{Rublee}, V.~{Rabaud}, K.~{Konolige}, and G.~{Bradski}, ``Orb: An efficient
  alternative to sift or surf,'' in \emph{2011 International Conference on
  Computer Vision}, Nov 2011, pp. 2564--2571.

\bibitem{Krig2016}
\BIBentryALTinterwordspacing
S.~Krig, \emph{Interest Point Detector and Feature Descriptor Survey}.\hskip
  1em plus 0.5em minus 0.4em\relax Cham: Springer International Publishing,
  2016, pp. 187--246. [Online]. Available:
  \url{https://doi.org/10.1007/978-3-319-33762-3_6}
\BIBentrySTDinterwordspacing

\bibitem{pandey2015passive}
R.~C. Pandey, R.~Agrawal, S.~K. Singh, and K.~K. Shukla, ``Passive copy move
  forgery detection using surf, hog and sift features,'' in \emph{Proceedings
  of the 3rd International Conference on Frontiers of Intelligent Computing:
  Theory and Applications (FICTA) 2014}.\hskip 1em plus 0.5em minus 0.4em\relax
  Springer, 2015, pp. 659--666.

\bibitem{kortli}
\BIBentryALTinterwordspacing
Y.~Kortli, M.~Jridi, A.~Al~Falou, and M.~Atri, ``A comparative study of cfs,
  lbp, hog, sift, surf, and brief for security and face recognition,'' in
  \emph{Advanced Secure Optical Image Processing for Communications}, ser.
  2053-2563.\hskip 1em plus 0.5em minus 0.4em\relax IOP Publishing, 2018, pp.
  13--1 to 13--22. [Online]. Available:
  \url{http://dx.doi.org/10.1088/978-0-7503-1457-2ch13}
\BIBentrySTDinterwordspacing

\bibitem{ali2016}
N.~Ali, K.~B. Bajwa, R.~Sablatnig, S.~A. Chatzichristofis, Z.~Iqbal, M.~Rashid,
  and H.~A. Habib, ``A novel image retrieval based on visual words integration
  of sift and surf,'' \emph{PloS one}, vol.~11, no.~6, p. e0157428, 2016.

\bibitem{monzo2011precise}
D.~Monzo, A.~Albiol, J.~Sastre, and A.~Albiol, ``Precise eye localization using
  hog descriptors,'' \emph{Machine Vision and Applications}, vol.~22, no.~3,
  pp. 471--480, 2011.

\bibitem{feature_survey_2018}
S.~A.~K. {Tareen} and Z.~{Saleem}, ``A comparative analysis of sift, surf,
  kaze, akaze, orb, and brisk,'' in \emph{2018 International Conference on
  Computing, Mathematics and Engineering Technologies (iCoMET)}, March 2018,
  pp. 1--10.

\bibitem{pan2009survey}
S.~J. Pan and Q.~Yang, ``A survey on transfer learning,'' \emph{IEEE
  Transactions on knowledge and data engineering}, vol.~22, no.~10, pp.
  1345--1359, 2009.

\bibitem{zheng2017sift}
L.~Zheng, Y.~Yang, and Q.~Tian, ``Sift meets cnn: A decade survey of instance
  retrieval,'' \emph{IEEE transactions on pattern analysis and machine
  intelligence}, vol.~40, no.~5, pp. 1224--1244, 2017.

\bibitem{alex_2012}
\BIBentryALTinterwordspacing
A.~Krizhevsky, I.~Sutskever, and G.~E. Hinton, ``Imagenet classification with
  deep convolutional neural networks,'' in \emph{Advances in Neural Information
  Processing Systems 25}, F.~Pereira, C.~J.~C. Burges, L.~Bottou, and K.~Q.
  Weinberger, Eds.\hskip 1em plus 0.5em minus 0.4em\relax Curran Associates,
  Inc., 2012, pp. 1097--1105. [Online]. Available:
  \url{http://papers.nips.cc/paper/4824-imagenet-classification-with-deep-convolutional-neural-networks.pdf}
\BIBentrySTDinterwordspacing

\bibitem{Wang_2020}
\BIBentryALTinterwordspacing
Y.~Wang, Q.~Yao, J.~T. Kwok, and L.~M. Ni, ``Generalizing from a few examples:
  A survey on few-shot learning,'' \emph{ACM Comput. Surv.}, vol.~53, no.~3,
  Jun. 2020. [Online]. Available: \url{https://doi.org/10.1145/3386252}
\BIBentrySTDinterwordspacing

\bibitem{bolme2010visual}
D.~Bolme, ``Visual object tracking using adaptive correlation filters,'' in
  \emph{The IEEE Conference on Computer Vision and Pattern Recognition CVPR)
  Year}, 2010.

\bibitem{Danelljan2014}
M.~Danelljan, G.~Häger, and F.~Khan, ``Accurate scale estimation for robust
  visual tracking,'' \emph{British Machine Vision Conference}, pp. 1--11, 01
  2014.

\bibitem{Danelljan2015}
M.~{Danelljan}, G.~{Häger}, F.~S. {Khan}, and M.~{Felsberg}, ``Learning
  spatially regularized correlation filters for visual tracking,'' in
  \emph{2015 IEEE International Conference on Computer Vision (ICCV)}, Dec
  2015, pp. 4310--4318.

\bibitem{Danelljan2016}
------, ``Discriminative scale space tracking,'' \emph{IEEE Transactions on
  Pattern Analysis and Machine Intelligence}, vol.~39, no.~8, pp. 1561--1575,
  Aug 2017.

\bibitem{Danelljan2016_2}
M.~Danelljan, A.~Robinson, F.~S. Khan, and M.~Felsberg, ``Beyond correlation
  filters: Learning continuous convolution operators for visual tracking,'' in
  \emph{European Conference on Computer Vision}.\hskip 1em plus 0.5em minus
  0.4em\relax Springer, 2016, pp. 472--488.

\bibitem{Danelljan2017eco}
M.~Danelljan, G.~Bhat, F.~Shahbaz~Khan, and M.~Felsberg, ``Eco: Efficient
  convolution operators for tracking,'' in \emph{Proceedings of the IEEE
  conference on computer vision and pattern recognition}, 2017, pp. 6638--6646.

\bibitem{donahue14}
\BIBentryALTinterwordspacing
J.~Donahue, Y.~Jia, O.~Vinyals, J.~Hoffman, N.~Zhang, E.~Tzeng, and T.~Darrell,
  ``Decaf: A deep convolutional activation feature for generic visual
  recognition,'' in \emph{Proceedings of the 31st International Conference on
  Machine Learning}, ser. Proceedings of Machine Learning Research, E.~P. Xing
  and T.~Jebara, Eds., vol.~32, no.~1.\hskip 1em plus 0.5em minus 0.4em\relax
  Bejing, China: PMLR, 22--24 Jun 2014, pp. 647--655. [Online]. Available:
  \url{http://proceedings.mlr.press/v32/donahue14.html}
\BIBentrySTDinterwordspacing

\bibitem{Snell2017}
\BIBentryALTinterwordspacing
J.~Snell, K.~Swersky, and R.~Zemel, ``Prototypical networks for few-shot
  learning,'' pp. 4077--4087, 2017. [Online]. Available:
  \url{http://papers.nips.cc/paper/6996-prototypical-networks-for-few-shot-learning.pdf}
\BIBentrySTDinterwordspacing

\bibitem{Mateen2019}
M.~Mateen, J.~Wen, Nasrullah, S.~Song, and Z.~Huang, ``Fundus image
  classification using vgg-19 architecture with pca and svd,'' \emph{Symmetry},
  vol.~11, p.~1, 2019.

\bibitem{RM2019}
F.~Özyurt, ``Efficient deep feature selection for remote sensing image
  recognition with fused deep learning architectures,'' \emph{The Journal of
  Supercomputing}, pp. 1--19, 12 2019.

\bibitem{Xu_2017}
\BIBentryALTinterwordspacing
Y.~Xu, Z.~Jia, L.-B. Wang, Y.~Ai, F.~Zhang, M.~Lai, and E.~I.-C. Chang, ``Large
  scale tissue histopathology image classification, segmentation, and
  visualization via deep convolutional activation features,'' \emph{BMC
  Bioinformatics}, vol.~40, May 2017. [Online]. Available:
  \url{https://doi.org/10.1186/s12859-017-1685-x}
\BIBentrySTDinterwordspacing

\bibitem{Liu_2017}
Q.~{Liu}, R.~{Hang}, H.~{Song}, and Z.~{Li}, ``Learning multiscale deep
  features for high-resolution satellite image scene classification,''
  \emph{IEEE Transactions on Geoscience and Remote Sensing}, vol.~56, no.~1,
  pp. 117--126, Jan 2018.

\bibitem{Ma_2015_ICCV}
C.~Ma, J.-B. Huang, X.~Yang, and M.-H. Yang, ``Hierarchical convolutional
  features for visual tracking,'' in \emph{The IEEE International Conference on
  Computer Vision (ICCV)}, December 2015.

\bibitem{wang2018face}
M.~Wang and W.~Deng, ``Deep face recognition: A survey,'' 2018.

\bibitem{wang-2018}
------, ``Deep visual domain adaptation: A survey,'' \emph{Neurocomputing},
  vol. 312, pp. 135--153, 2018.

\bibitem{Mcneely2020}
D.~McNeely-White, J.~R. Beveridge, and B.~A. Draper, ``Inception and resnet
  features are (almost) equivalent,'' \emph{Cognitive Systems Research},
  vol.~59, pp. 312--318, 2020.

\bibitem{cohen_2020}
U.~Cohen, S.~Chung, D.~D. Lee, and H.~Sompolinsky, ``Separability and geometry
  of object manifolds in deep neural networks,'' \emph{Nature Communications},
  vol.~11, no.~1, 2020.

\bibitem{mohapatra_2014}
D.~P. Mohapatra and S.~Patnaik, \emph{Intelligent Computing, Networking, and
  Informatics Proceedings of the International Conference on Advanced
  Computing, Networking, and Informatics, India, June 2013}.\hskip 1em plus
  0.5em minus 0.4em\relax Springer India, 2014.

\end{thebibliography}

\end{document}